\documentclass[conference]{IEEEtran}
\IEEEoverridecommandlockouts
\usepackage{cite}
\usepackage{amsmath,amssymb,amsfonts}
\usepackage{algorithmic}
\usepackage{graphicx}
\usepackage{textcomp}
\usepackage{xcolor}
\def\BibTeX{{\rm B\kern-.05em{\sc i\kern-.025em b}\kern-.08em
    T\kern-.1667em\lower.7ex\hbox{E}\kern-.125emX}}
\begin{document}

\title{Identifying the Context Shift between Test Benchmarks and Production Data\\
\thanks{The author would like to acknowledge funding from the MIT Media Lab member companies}
}

\author{\IEEEauthorblockN{Matthew Groh}
\IEEEauthorblockA{\textit{MIT Media Lab} \\
Cambridge, MA \\
groh@mit.edu}
}

\maketitle

\begin{abstract}
Machine learning models are often brittle on production data despite achieving high accuracy on benchmark datasets. Benchmark datasets have traditionally served dual purposes: first, benchmarks offer a standard on which machine learning researchers can compare different methods, and second, benchmarks provide a model, albeit imperfect, of the real world. The incompleteness of test benchmarks (and the data upon which models are trained) hinder robustness in machine learning, enable shortcut learning, and leave models systematically prone to err on out-of-distribution and adversarially perturbed data. The mismatch between a single static benchmark dataset and a production dataset has traditionally been described as a dataset shift (or distribution shift with subcategories including covariate shift, prior probability shift, and concept shift). These shifts are simultaneously over-specified with formal definitions for comparing two data samples and under-specified for evaluating the data-generating process that drives the mismatch between data samples. In an effort to clarify how to address the mismatch between test benchmarks and production data, we introduce context shift to describe semantically meaningful changes in the underlying data generation process. Moreover, we identify three methods for addressing context shift that would otherwise lead to model prediction errors: first, we describe how human intuition and expert knowledge can identify semantically meaningful features upon which models systematically fail, second, we detail how dynamic benchmarking – with its focus on capturing the data generation process – can promote generalizability through corroboration, and third, we highlight that clarifying a model's limitations can reduce unexpected errors. Robust machine learning is focused on model performance beyond benchmarks, and as such, we consider three model organism domains – facial expression recognition, deepfake detection, and medical diagnosis – to highlight how implicit assumptions in benchmark tasks lead to errors in practice. By paying close attention to the role of context in a prediction task, researchers can design more comprehensive benchmarks, reduce context shift errors, and increase generalization performance.

\end{abstract}


\section{Machine Learning Models are Brittle in Production}

Dataset benchmarks offer a standard for comparing and evaluating the performance of machine learning models on real-world tasks like object detection~\cite{deng2009imagenet}, handwritten digit recognition~\cite{deng2012mnist}, image captioning~\cite{chen2015microsoft}, general language understanding~\cite{wang2018glue}, affect recognition~\cite{kosti_context_2019}, deepfake detection~\cite{dolhansky_deepfake_2020}, medical diagnosis (e.g. for skin disease~\cite{daneshjou2022disparities}, pneumonia~\cite{irvin2019chexpert}, critical care~\cite{johnson2016mimic}, etc.), and many other tasks. 

As a standard for comparison, dataset benchmarks have enabled rapid progress in computer vision and natural language processing. However, despite intentions to create and curate data that match the real-world as closely as possible, the dynamic, high-dimensional, combinatoric complexity of many real-world tasks can be difficult to fully capture in a single static benchmark. Indeed, the development and evaluation of machine learning models on benchmarks often suffer from a variety of historical, representational, measurement, aggregation, and evaluation biases which can be further exacerbated by deployment biases where the task that a benchmark is intended to measure differs from the real-world task~\cite{suresh2021framework, srinivasan2021biases}. Moreover, data for benchmarks are often collected at scale~\cite{russakovsky2015imagenet}, which leaves data open to poisoning attacks~\cite{chen_targeted_2017}, leakage~\cite{kapoor2022leakage}, subjectivity and multiple interpretations~\cite{gordon_jury_2022} and error~\cite{northcutt_pervasive_2021}. As a consequence, machine learning models that appear to be approaching (and sometimes surpassing) human-level ability on a test benchmark will often error when shown out-of-distribution~\cite{torralba2011unbiased} data. In other words, the reliance on static test benchmarks as metrics for projecting production performance~\cite{thomas2022reliance} inflates the accuracy of machine learning model performance and leaves open the questions, ``Can you trust your model? Will it work in deployment?''~\cite{lipton_mythos_2018}

The meaning of out-of-distribution data depends on a task's context. One classic example of out-of-distribution data in object detection tasks from the invariant risk minimization literature is an image of a cow on a sandy beach (and similarly a camel on a green pasture) because training data rarely contains such animal-environment pairs~\cite{arjovsky_invariant_2020}. Such a lack of uncommon animal-environment pairs leads to machine learning shortcuts where a model learns spurious correlations (e.g. cloven hoofed mammals next to sand are camels and the ones next to grass are cows). With \textit{a priori} knowledge of likely model shortcuts, one approach for addressing this kind of model brittleness is to include auxiliary labels that can serve as a causally-motivated regularization framework~\cite{makar2022causally}. However, for identifying previously unknown shortcuts, post hoc model explanations are often ineffective ~\cite{adebayo2021post} (though explanations via concept traversals~\cite{ghandeharioun2021dissect} and identifying model failures as directions in latent space via contrastive learning where images and natural language are embedded in a shared latent space show promise~\cite{jain2022distilling}). In contrast, human intuition can identify many out-of-distribution contexts on which shortcut learning may occur.

In one of the clearest examples of the benchmark-production gap, researchers closely followed the original dataset generation processes to build new test sets for two of the most commonly used dataset benchmarks for object recognition, ImageNet~\cite{deng2009imagenet} and CIFAR-10~\cite{krizhevsky2014cifar}, and found significant drops (ranging from 3\% to 15\% in 67 models applied to ImageNet and 34 models applied to CIFAR-10) in the state-of-the-art models' performance from the original benchmarks to the newly recreated benchmarks. This drop in performance does not appeared to be explained by random sampling error, hyperparameter tuning for optimizing performance on the original test set, or obvious changes in semantically meaningful features, but instead the performance gap appears to arise from subtle changes in the data~\cite{recht2019imagenet}. Although object recognition may seem like a straightforward task, it involves many tacit contexts. 

In complex human-centered machine learning applications, a task's context involves answers to the following kinds of questions: What is the task? For whom is the task designed? When and where does it take place? Why is it done? Are there any interventions happening that might alter features and labels associated with the task? And how is the task measured? The lack of clear answers to these questions indicates that the model and its evaluation lack generalizability simply because it is not clear what the model should generalize. Likewise, clear answers to these questions without a corresponding diverse representation in the benchmark dataset to evaluate performance along these contexts leave the model prone to generalization errors in out-of-distribution contexts. 

As an example of a human-centered machine learning generalization failure, consider facial recognition. In Joy Buolamwini's and Timnit Gebru's algorithmic audit of facial recognition benchmarks and classifiers, they reveal the most commonly used benchmarks were composed of images of people with predominantly light skin – in other words, images of people with dark skin were relatively out-of-distribution~\cite{buolamwini2018gender}. Furthermore, the audit presented a new benchmark to evaluate accuracy across intersectional identities, and commercial gender classification models performed extremely accurately in identifying men with light skin (with a maximum error rate of less than 1\%) but incorrectly in women with dark skin (with a maximum error rate of 35\%)~\cite{buolamwini2018gender}. This large accuracy disparity reveals how failures to generalize can be hidden by benchmarks that do not represent the diversity of the real world. Research on machine learning applied to the diagnosis of skin disease reveals a similar story to facial recognition: models trained to classify skin disease based on images of only light or dark skin are more accurate in skin tones closest to the skin tones in the images in which the model was trained~\cite{groh_evaluating_2021}. These examples corroborate the notion that simply optimizing for predictive accuracy with very large datasets can often misrepresent the true data generating process and lead to systematic errors~\cite{hullman2022worst}.

In other domains like affect recognition, an out-of-distribution context can be very task specific; for example, spontaneous facial expressions can be out-of-distribution for facial expression benchmarks that primarily contain posed expressions~\cite{dupre_performance_2020} and emotion categories like anger or surprise can be out-of-distribution for the same benchmarks where happy and neutral labels are most common~\cite{li_deep_2020}. 

Adversarial perturbations – a special case of out-of-distribution data - offer a systematic method for identifying models' failures to generalize. Examples of adversarial perturbations include a small sticker on a stop sign~\cite{brown2017adversarial, eykholt2018robust}, a mainly translucent sticker on the lens of a camera~\cite{li2019adversarial}, noise or rotations in medical images, and text substitution in medical notes and reimbursement codes~\cite{finlayson2019adversarial} that can fool machine learning models into picking up patterns from adversarial perturbations without changing the semantically meaningful features apparent to humans. These adversarial perturbations demonstrate a lack of model robustness~\cite{ilyas2019adversarial}, lead to model errors that reasonable humans would rarely make, and open the question: How can we build models that are invariant to the same semantically meaningful features to which humans are invariant? Training robust models with adversarial perturbations is a starting point that tends to align more closely with human perceptions than standard training~\cite{tsipras2018robustness}, but it does not directly address semantically meaningful features that are only indirectly considered when examining distribution shifts.

\section{Systematic Errors Arise from Context Shift Which Drives Distribution Shift}

\begin{figure}[htbp]
\centerline{\includegraphics[width=.49\textwidth]{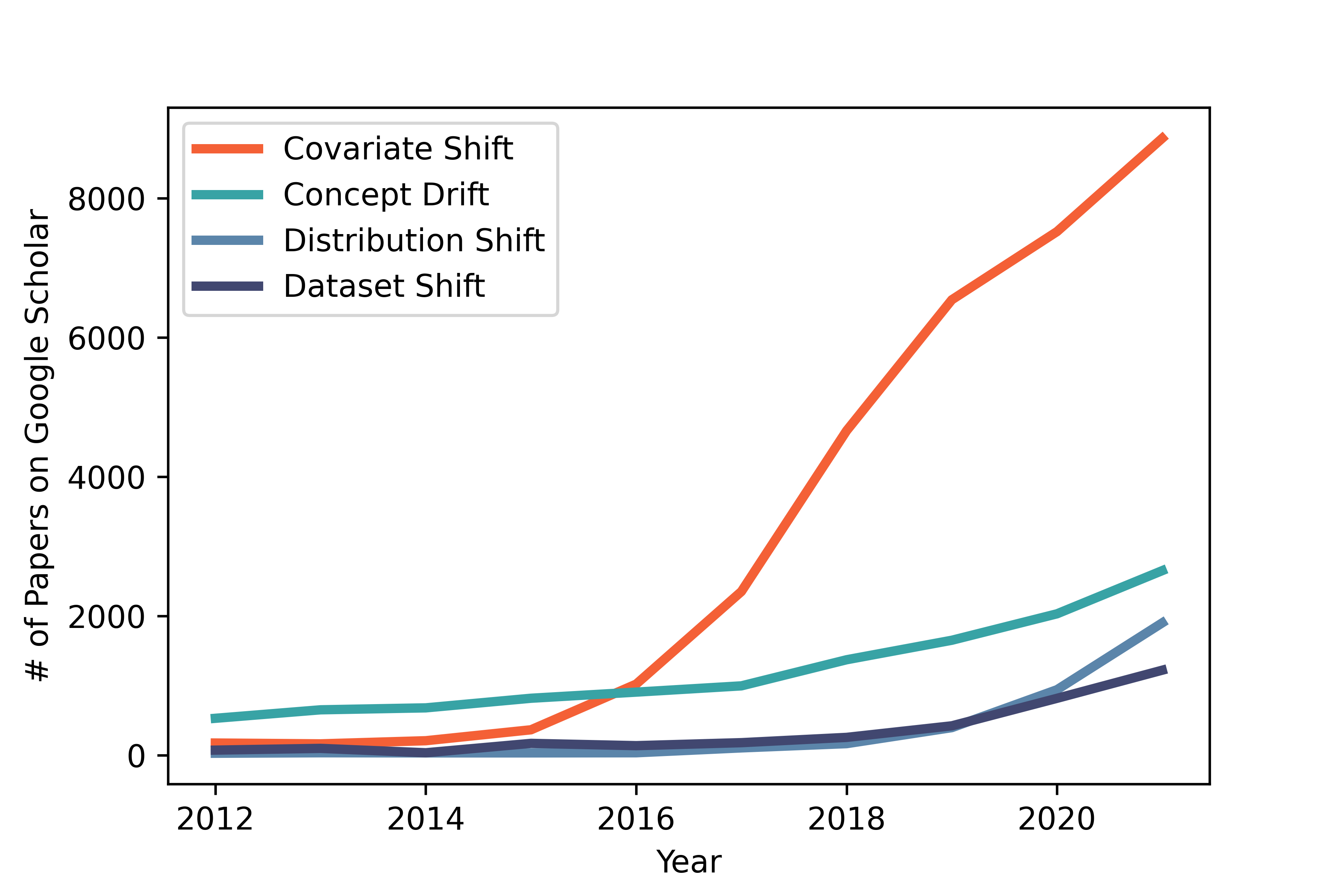}}
\caption{Number of papers on Google Scholar from 2012 to 2021 for search queries combining ``machine learning'' + the four most common terms for distribution shifts. For context, ``machine learning'' returns 185,000 articles in 2012 and 245,000 articles in 2021. The terms ``prior probability shift'' and ``concept shift'' return 398 and 1,040 papers over all time, respectively, when paired with ``machine learning''.}
\label{fig:google_scholar_results}
\end{figure}

The mismatch between two datasets (that is, a test benchmark and a sample drawn from production data) has been traditionally described as a dataset shift~\cite{quinonero2008dataset} and recently has been described more frequently as a distribution shift. In Figure~\ref{fig:google_scholar_results}, we present the number of papers on Google Scholar containing both ``machine learning'' and ``distribution shift'' (and other shift keywords). 

\begin{figure*}[htbp]
\centerline{\includegraphics[width=.75\textwidth]{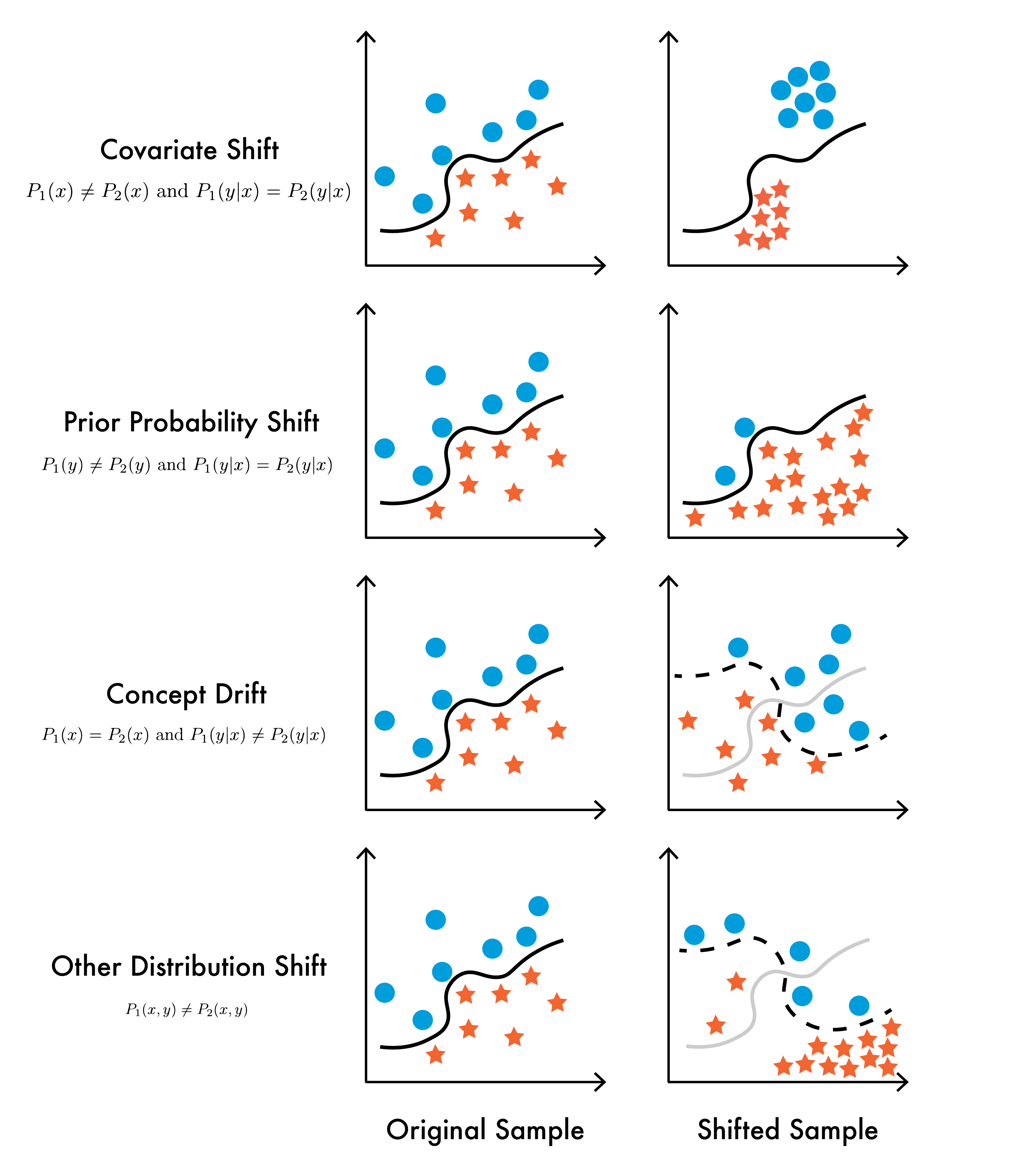}}
\caption{Illustrations of the four kinds of distribution shifts (defined in Moreno et al. 2012~\cite{moreno2012unifying}) with spatial positions representing the feature space, geometric shapes and colors representing the ground truth label, the solid boundary line representing the learned representation of labels from the original sample, and the dotted boundary line representing the learned representation of labels from the shifted sample. Most real-world distribution shifts involve changes across features, labels, and the relationship between features and labels, and as such would be characterized as ``Other Distribution Shift.'' The core problem with the conceptual framework of distribution shift is that it is merely a symptom of changes in data-generating processes - how data are created, collected, and curated – but not part of the data-generating process itself, which needs to be directly addressed to improve model reliability and robustness.    }
\label{fig:distribution_shift}
\end{figure*}

Dataset shift and distribution shift both refer to the non-equivalence of the joint distributions between two datasets; formally, distribution shift describes the following equation $P_1(y,x) \neq P_2(y,x)$ where $P_n(y,x)$ is the joint distribution of labels, $y$, and covariates, $x$ for a particular dataset, $n$~\cite{moreno2012unifying}. Based on Moreno et al 2012, the subcategories of distribution shift include covariate shift when the distribution of features changes but everything else remains the same, prior probability shift when the distribution of labels changes but everything else remains the same, and concept shift (more commonly referred to as concept drift) when the distribution of labels conditional on features changes but everything else remains the same. We illustrate examples of each shift in Figure~\ref{fig:distribution_shift} to motivate intuition as to how the changes appear. These distribution shifts have been formally specified as follows~\cite{moreno2012unifying}:

\begin{itemize}
  \item Covariate shift: $P_1(x) \neq P_2(x)$ but $P_1(y|x)=P_2(y|x)$
  \item Prior probability shift: $P_1(y) \neq P_2(y)$ but $P_1(y|x)=P_2(y|x)$
  \item Concept drift: $P_1(y|x) \neq P_2(y|x)$ but $P_1(x) = P_2(x)$
  \item Other distribution shift: $P_1(y,x) \neq P_2(y,x)$ where none of the above three shifts applies.
\end{itemize}

Although covariate shift, prior probability shift, and concept drift can be formally specified, empirically evaluated, and in some specific cases mitigated, there has not been and will not be a perfect end-to-end solution for generalizing from a benchmark dataset that is not an independent and identically distributed (i.i.d.) sample of a data generating process. In the language of causal representation learning~\cite{scholkopf2021toward}, data are often assumed to be observational (sampled from an unchanging distribution), but in the real world it is often interventional (sampled from a changing distribution), and interventions can be particularly difficult to directly learn from the data when data are raw (perceptual inputs like images and audio) as opposed to hand-crafted (semantically meaningful features) because of the combinatorial explosion of potential contexts. The core problem with the conceptual framing of distribution shift is that it is merely a symptom of changes in data-generating processes - how data are created, collected, and curated – but not part of the data-generating process itself, which needs to be directly addressed for improving model reliability and robustness.

Context refers to the semantically meaningful upstream factors that drive the distribution shift. Machine learning research has long considered the hidden contexts behind the distribution shift~\cite{widmer1996learning}. In this paper, we seek to clarify the overlapping and often confusing jargon of distribution shift by introducing context shift as a term to describe the semantically meaningful features that influence data-generating processes and focusing on how to identify the changes in the creation, collection, and curation of data that lead to distribution shifts. By centering the problem of robustness and generalizability of applied machine learning on context shift, we expand and orient the problem space to data-centered human intuition and expertise. 

\begin{figure*}[htp]
\centerline{\includegraphics[width=.75\textwidth]{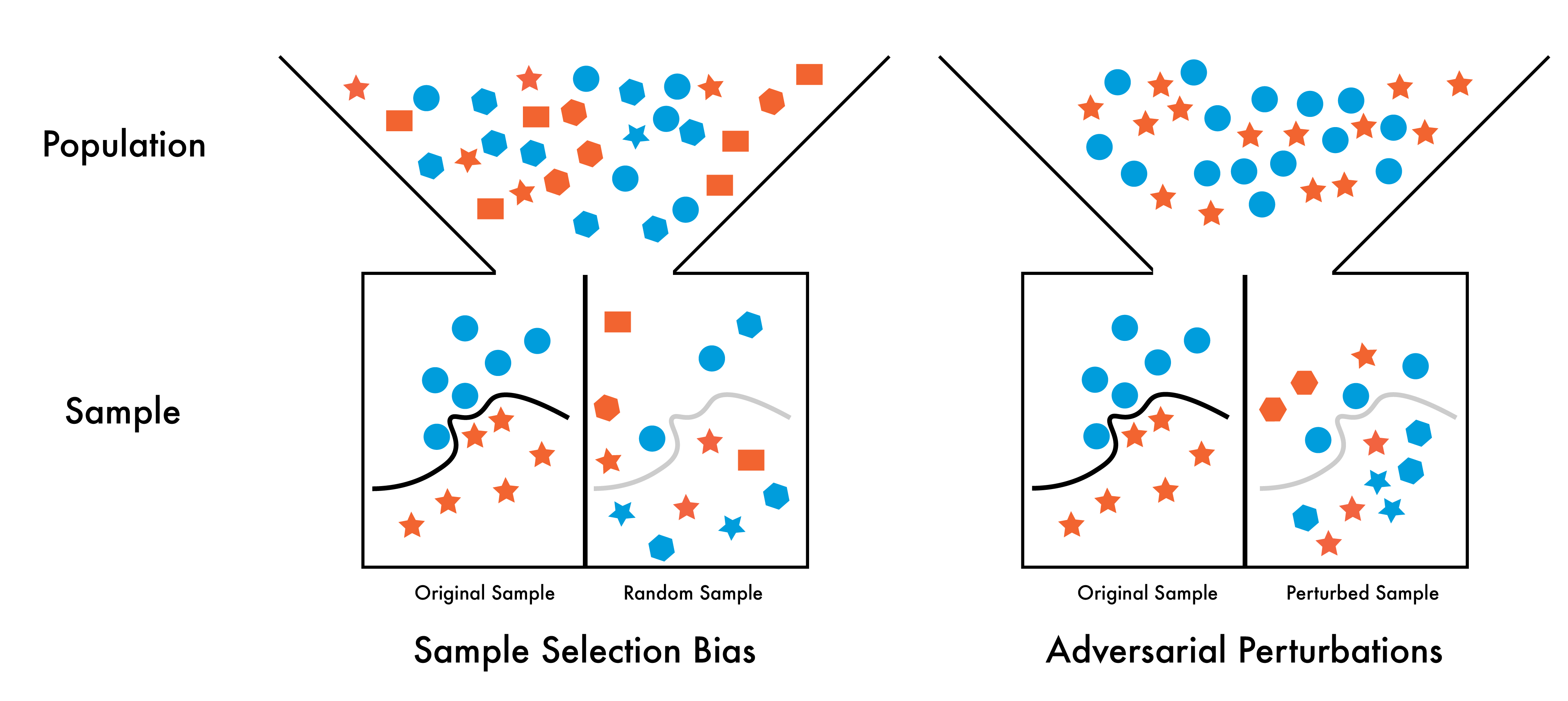}}
\caption{Illustrations of sample selection bias and adversarial perturbations with colors representing the ground truth label, geometric shapes and spatial positions representing the features, the top of the funnel representing the full populations, the bottom of the funnel representing the samples drawn from the population, and the solid boundary line representing the learned representation of labels from the original sample. On the left, the population contains upright stars, rotated stars, hexagons, rectangles, and circles, but the biased original sample only contains circles and stars. The random sample contains much higher diversity of features and relationships between features and labels. As such, the learned representation fails in more than 50\% of observations. On the right, the population contains upright stars and blue circles. The original sample contains the same set of features, but the perturbed sample includes both rotated hexagons and stars, which may not be immediately noticeable to humans at first glance. Depending on the rotation, the learned representation misclassifies the perturbed shapes. Both pairs of samples present changes in features and changes in labels conditional on the features, which would make these examples of ``Other Distribution Shift.'' This figure is intended to provide intuition for where the perspective of distribution shift is inadequate and where the perspective of identifying semantically meaningful features that influence how samples are curated and created can inform approaches for addressing robustness in applications of machine learning. }
\label{fig:context_shift}
\end{figure*}

Rather than focusing on differences in two distributions with disregard for the reasons behind the difference, we suggest researchers consider three concepts that drives semantically meaningful changes in datasets: sample selection bias (e.g. the new dataset contains images of people from a demographic not represented in the old dataset), adversarial perturbations (e.g. the new dataset contains noise injections that are imperceptible to the human eye but change model performance), or non-stationarity (e.g. the new dataset contains images of smart phones post 2018 but the old dataset only contains flip phones before 2010). While we list non-stationarity separately from sample selection bias, non-stationarity can be considered a special case of sample selection bias where the sample selection bias is over features and labels which may appear different in the future. We present Figure~\ref{fig:context_shift} to illustrate two examples of context shift.

Unlike distribution shift, which can be measured between two datasets, context shift can only be fully addressed by learning the entire population's data distribution, the kinds of changes that are and are not perceptible to humans, and how the population's data distribution changes over time and space. Outside of artificially constrained spaces like synthetic datasets or games, access to the entire populations data distribution (or the rules governing the distribution) across space and time is rare. Nevertheless, people generally have intuition and the ability to reason about data distributions of combinatoric contexts that they might never experience. In fact, cognitive science research shows that intuitive reasoning about statistical power analysis begins early in childhood~\cite{pelz2022foundations}.

\section{Addressing Robustness with Human Intuition and Expertise}

Over the last few years, researchers have been developing data-centered frameworks to offer guidance for breaking down the data generating process into relevant component parts that reveal where context shift may lead to benchmark-production performance gaps. These frameworks include \textit{Data Statements for Natural Language Processing}~\cite{bender2018data}, \textit{The Dataset Nutrition Label}~\cite{holland2018dataset}, \textit{Model Cards for Model Reporting}~\cite{mitchell2019model}, \textit{Datasheets for Datasets}~\cite{gebru2021datasheets}, \textit{Closing the AI accountability gap}~\cite{raji_ai_2021}, \textit{The Ethical Pipeline for Healthcare Model Development}~\cite{chen_ethical_2020}, and \textit{The Clinician and Dataset Shift in Artificial Intelligence}~\cite{finlayson_clinician_2021}. Likewise, meta-frameworks offer guidance for ensuring data documentation frameworks are useful and actionable~\cite{heger2022understanding}. 

As a heuristic for human-centered machine learning applications, teams of conscientious, creative, and skilled model developers, data engineers, and subject matter experts may find it useful to identify a first-order, non-exhaustive list of dimensions on which context shift is likely to occur. This list of dimensions depends largely on the context and the degree to which the data are subjective, representative, and missing~\cite{mullainathan_does_2017}. In ethnographic interviews with machine learning engineers, researchers find that engineers often address changes in context with ``elaborate rule-based guardrails to avoid incorrect outputs''~\cite{shankar2022}. Recent examples of semantically meaningful dimensions that have been demonstrated as useful for evaluating robustness in applied machine learning include skin color in face recognition~\cite{buolamwini2018gender} and dermatology diagnosis~\cite{groh_evaluating_2021, daneshjou2022disparities}, background scenery for affect recognition~\cite{kosti_context_2019}, number of people in a video for deepfake detection~\cite{groh_deepfake_2022}, number of chronic illnesses for algorithmic healthcare risk prediction~\cite{obermeyer_dissecting_2019}, data artifacts like surgical markings~\cite{winkler_association_2019} or clinically irrelevant labels~\cite{oakden-rayner_hidden_2020} for medical diagnosis classification, patients' self reports of pain for quantifying severity of knee osteoarthritis~\cite{pierson_algorithmic_2021}, and image similarity characteristics for pathologists to disambiguate between machine learning and user errors~\cite{cai_human-centered_2019}. 

Knowledge elicitation is not a solved problem, but helpful questions that may guide the identification of potential context shifts in complex, human-centered machine learning applications include (and are not limited to): who are represented in the data and as annotators of the data, when and where is the data collected, how do social, geographical, temporal, technological, aesthetic, financial incentives and other idiosyncrasies influence the creation of the data, and why the data is curated as it is. Knowledge elicitation has been historically ill-defined in artificial intelligence applications~\cite{forsythe1993engineering}, but recent work developing taxonomies for knowledge elicitation helps to formalize the process and increase transparency along the way~\cite{kerrigan2021survey, chen2022perspectives}.

Another expert intuition guided approach to closing the benchmark-production gap involves developing test benchmarks with adequate diversity in the data along the contextual dimensions upon which human intuition and expertise suggests model performance is most likely to vary. Recent examples of benchmark datasets working towards this goal are \textit{BREEDS: Benchmarks for Subpopulation Shift}~\cite{santurkar_breeds_2020} and \textit{WILDS: A Benchmark of in-the-Wild Distribution Shifts}~\cite{koh_wilds_2021}, which includes labels for relative contexts and sub-populations for the explicit examination of context shifts.

\section{Addressing Robustness with Dynamic Benchmarking}

A second approach to addressing the benchmark-production gap is to transform the practice of evaluation from static benchmarks to dynamic benchmarks where models' performance is not evaluated on a single dataset, but rather continually evaluated on datasets produced via well-specified, quality controlled data generation processes. One example of dynamic benchmarking is dynabench~\cite{kiela2021dynabench}, which is designed for natural language processing tasks. For general development of dynamic benchmarks, data generation process desiderata should include specifying the following dimensions of a dynamic benchmark: 

\begin{itemize}
  \item \textbf{Prediction task}: What are the input features and output labels? For example, inputs may be images and outputs may be lists of objects or inputs may be described more specifically as images of skin lesions photographed by dermatoscopes and outputs may be classifications of benign and malignant by board-certified dermatologists in the United States. It is important to be careful that the task matches the expected goal because unexpected mismatches between tasks and goals are relatively common~\cite{mullainathan_inequity_2021, kerr1975folly}.
  
  \item \textbf{Ground truth annotation arbitration}: Who has the authority to annotate the data? How do experts differ from crowdworkers or an algorithm~\cite{groh2022towards}? How should the data be annotated? How should inter-annotator disagreement be represented? What categories should be included? 
  
  \item \textbf{Data inclusion and exclusion criteria}: What are the possible data sources? How are data curated from these sources? What is the data distribution of categories and subcategories? What are the quality constraints?
  
  \item \textbf{Benchmark size and shape}: What is the minimum size of a batch of data to serve as a benchmark? How should benchmarks by different groups for the same task be combined together?
  
 \end{itemize}
 
These desiderata enable the development of dynamic benchmarks that further enable quantitative evaluation of model robustness via corroborated accuracy, which is the distribution of accuracy scores across dynamic benchmarks. Rather than simply evaluating a model on a single or a few static test benchmarks, we might consider a well-corroborated model to be one that meets two criteria: first, it is reasonably available for evaluation, and second, all attempts to uncover systematic errors in well-specified contexts reveal no significant accuracy disparities. The practice of dynamic benchmarking could be particularly relevant for addressing the \textit{AI Knowledge Gap}~\cite{epstein_closing_2018} characterized by the disparity between the large number of machine learning models and the small number of studies evaluating these models' performance. Furthermore, dynamic benchmarking can be combined with benchmark task misalignment methodologies~\cite{tsipras2020imagenet, ilyas2022datamodels} to assess how aligned (or misaligned) model predictions are with human annotations and considering diverse examples that bring transparency to the ethical implications and societal impact to model development~\cite{paullada2021data}.

The transition from static benchmarks on a particular instance (or set of instances) to dynamic benchmarks on data generation processes defined by explicit desiderate may be useful for addressing the fundamental issue of construct validity that arises in singular, static benchmarks~\cite{raji_ai_2021}.

\section{Addressing Robustness by Clarifying a Model's Limitations}

A third approach to reducing the benchmark-production gap is to appropriately specify the contexts in which a model is expected to work via a limitations section~\cite{smith2022real}. 

To clarify domain-specific limitations driving the benchmark-production gap, we consider implicit assumptions that lead to a context shift in three real-world computer vision tasks: facial expression recognition, deepfake detection, and medical diagnosis. 

\section{Model Organism Domains for Addressing Context Shift in Applied Machine Learning}

\subsection{Facial Expression Recognition}

In the field of affective computing, facial expression recognition (FER) is a task to classify human facial expressions with affective labels~\cite{cohn2015automated, li_deep_2020}, which can be a useful component in designing human-AI interactions with computational empathy~\cite{picard2000affective, paiva2017empathy, groh2022computational}. Model-based FER is similar to how humans recognize the emotions of others (called empathic accuracy in affective science~\cite{ickes1993empathic} and emotion reasoning in developmental psychology~\cite{ruba_development_2020}) except that FER is based solely on facial expressions, whereas affect recognition can include information about someone's gestures, language, tone, physiological measurements, and the long-tail of context, which can include factors such as the temperature outside, the social relationship between two individuals, what happened the day before, and more.

Consider an example from relatively recent research~\cite{mollahosseini2016going} where a standard neural network architecture, AlexNet~\cite{krizhevsky2012imagenet}, is trained on a large number of images of spontaneous and posed facial expressions to classify images into seven categories (anger, disgust, fear, happiness, sadness, surprise, and neutral) and achieves accuracy scores ranging from 48.6\% in SFEW~\cite{dhall2011static} to 56.0\% in MMI~\cite{pantic2005web} to 56.1\% in DISFA~\cite{mavadati2013disfa} to 61.1\% in FER2013~\cite{goodfellow2013challenges} to 77.4\% in FERA~\cite{valstar2011first,banziger2010introducing} to 92.2\% in CK +~\cite{lucey2010extended} to 94.8\% in MultiPie~\cite{gross2010multi}. While this model's accuracy is significantly better than random guessing, which would be 14.2\%, it varies dramatically depending on the chosen benchmark dataset. How should we interpret a performance gain of 21.9 percentage points on one dataset and an average performance gain of 3.5 percentage points on the other 6 datasets in an alternative network architecture? How should we interpret the model's ability to achieve higher accuracy scores than non-neural network methods on three of the seven benchmark datasets? What does the distribution of performance tell us about how this model would perform on real-world production data? There is no clear answer to any of these questions, yet an implicit assumption in the well-cited, peer-reviewed publication of this FER paper is the slightly improved performance on several benchmark datasets appears to mark a contribution to the field of facial expression recognition. This assumption has the potential to lead to another more pernicious and mistaken assumption: the role of contextual features for real-world performance can be ignored when assessing the state-of-the-art methodology in applied problems like FER. 

Clearly, models can learn facial expression features that map to human annotations of a handful of emotion categories to classify images at significantly better than chance rates. But, it is not reported nor clear how changes in lighting, head pose, occlusion, skin tone, ethnicity, age, gender, and background scenery influence both the model's performance or human annotations. It is also underexplored how well FER models would perform if humans of diverse cultures annotated these images. Likewise, it is unclear how the model would perform on more fine-grained emotion categories~\cite{cowen_sixteen_2021} or labels based on affective dimensions like valence, arousal, and dominance. Furthermore, in many real-world settings where people may feign smiles to appease their managers, cry to express joy, or appear neutral to hide a winning poker hand, the perspective of outside observers may be very different than the perspective of close friends or individuals themselves. We highlight these relevant contextual features to highlight the many dimensions in which context shift can occur between test benchmarks and real-world production data. While these are not an exhaustive list of contextual features, these represent intuitive, first-order contexts for conducting algorithmic audits, developing future benchmark datasets with these labeled contexts, and adapting models to handle these dimensions. While researchers build the next version of contextualized dynamic benchmarks, other researchers who are focused on developing models should at the very least include caveats in their papers about the likely contextual dimensions that may affect performance. 

\subsection{Deepfake Detection}

As a second case study of context shift in real-world applications of computer vision, we consider deepfake detection. Deepfakes are videos that have been manipulated to make someone appear to do or say something they have not said~\cite{boneh2020preparing}. These types of manipulation can be qualitatively characterized as face swapping where two people's faces are swapped, head puppetry where facial landmarks are adjusted to make someone appear to be speaking, and lip-syncing where an individual's lips are moved in sync with the phonemes from an external audio track~\cite{lyu_deepfake_2020}.

The largest deepfake detection benchmark dataset to date is the Deepfake Detection Competition Dataset (DFDC)~\cite{dolhansky_deepfake_2019, dolhansky_deepfake_2020}, which consists of 128,154 videos based on performances by 960 consenting actors representing diversity across sex and ethnicity. However, Groh et al 2022 point out,``Unlike viral deepfake videos of politicians and other famous people, the videos from [this benchmark dataset] have minimal context: These are all 10 [second] videos depicting unknown actors making uncontroversial statements in nondescript locations''~\cite{groh_deepfake_2022}. This deepfake test benchmark is designed to evaluate algorithmic performance in identifying videos that have (and have not) been manipulated by seven synthetic techniques. 

But, the real-world deepfake detection problem is not simply identifying whether one of seven synthetic techniques has been applied to a video. Instead, the real-world problem is identifying videos that have been algorithmically altered to impersonate innocent people and deceive the viewer. This problem is more than just a computer vision problem; it is a deception detection problem that involves both searching for artifacts that reveal that a manipulation has occurred and applying prior knowledge and critical reasoning to assess the likelihood that the video has been fabricated. 

The DFDC does not include politicians or any scenes of news conferences or people speaking to a large audience. If we assume that harmful deepfakes will involve these kinds of contexts (like a deepfake of President Volodomyr Zelensky that appeared in March 2022~\cite{wakefield_2022}), then it is important to evaluate models on videos with these kinds of dimensions, such as those from the Presidential Deepfakes Dataset~\cite{sankaranarayanan2021presidential,groh2022human} and the Protecting World Leaders against Deepfakes Dataset~\cite{agarwal2019protecting}. When Groh et al 2022 examined the leading state-of-the-art for detecting DFDC videos on deepfakes of Kim Jung-un and Vladimir Putin, they found the the leading model predicted a 2\% and 8\% likelihood these videos are deepfakes. While failure on two examples is only an anecdote, this failure speaks to an important need: diverse test benchmarks that cover the first-order dimensions where human intuition and expertise suggests context shift is most likely to occur. 

\subsection{Medical Diagnosis}

As a third case study of context shift, we consider medical diagnosis in store-and-forward teledermatology settings where clinical data are collected at one site and sent electronically for evaluation at another site. Recent research on machine learning applied to skin disease classification has demonstrated the human expert-level performance of models in a number of specific tasks~\cite{esteva_dermatologist-level_2017, liu_deep_2020}. However, it is unclear how these models will perform on people with dark skin because the first paper does not describe the distribution of ethnicity or skin tone in the evaluation benchmark~\cite{esteva_dermatologist-level_2017} and the evaluation benchmark in the second paper contains only 2.7\% of people with the second darkest of the six Fitzpatrick Skin Types (FSTs) and 1 person with the darkest of the FSTs~\cite{liu_deep_2020}. Given the accuracy disparities that appeared across skin types in facial recognition, expert intuition suggests that systematic errors are likely to also appear in skin disease classifiers. 

In fact, empirical research corroborates this intuition~\cite{groh_evaluating_2021}, and the Diverse Dermatology Images (DDI) dataset~\cite{daneshjou2022disparities} reveals that state-of-the-art skin disease classification models make systematically more errors on dark skin than on light skin. The DDI represents a more comprehensive benchmark than previous datasets, and as a result, the DDI exposed errors that should guide and motivate the future development of machine learning models towards more robustness. However, the DDI is not perfectly comprehensive; the dataset is de-identified for privacy reasons and lacks free text clinical notes and other information that physicians would acquire via an in-person examination~\cite{daneshjou2022disparities}. Given that many skin diseases appear similarly and expert diagnoses are based on clinical history and non-visual features, expert intuition would expect, once again, that systematic errors lurk in the state-of-the-art machine learning models for store-and-forward skin disease classification.

\section{Towards Robustness in Applied Machine Learning}

Supervised machine learning models are very good at identifying statistical regularities in a given dataset but tend to err on out-of-distribution data that may arise from sample selection bias, adversarial perturbation, or nonstationarity. On the other hand, humans can be quite good at identifying contextual examples of out-of-distribution data. By combining the strengths of machine learning models with human intuition and expertise, early career ancient historians can quickly restore and date ancient texts~\cite{assael_restoring_2022}, content moderation teams can more accurately distinguish between real and fake videos~\cite{groh_deepfake_2022}, and general practitioners can more accurately diagnose skin conditions from images~\cite{jain_development_2021} (although AI advice can also mislead experts; see~\cite{tschandl_humancomputer_2020, abeliuk_quantifying_2020, gaube_as_2021, jacobs_how_2021, vaccaro2019effects}). In fact, initial evidence suggests that human intuition is fairly accurate in predicting model misclassifications on common object detection tasks~\cite{zhou2020will}. The integration of machine predictions with human decisions in collaborative decision making systems may be the most immediately effective way to avoid errors from context shift.

Promising future research directions for developing robust machine learning models under distribution shift involve the following iterative process: first, identify missing contexts in test benchmarks, second, collect data that contain those missing contexts, and third, adjust the model accordingly. Researchers can begin to identify missing contexts by collaborating with human experts who may be able to identify first-order drivers of context shift on a task-by-task basis. Similarly, researchers can further identify missing contexts by evaluating models against data generation process desiderata rather than a single or a few datasets.

Finally, one of the most effective solutions for addressing the benchmark-production gap is for researchers to clearly communicate the contexts in which a model has been evaluated and the contexts in which the model's performance is unknown.

\section*{Acknowledgments}

I thank Katie Matton, Noah Jones, Robert Lewis, Rosalind Picard, Christelle Godin, Marion Mainsant, Adrien Viault, and Jessica Hullman for feedback on early drafts of this paper, and I thank Zach Lipton and other participants at the ICML Principles of Distribution Shift workshop for additional insightful feedback. 

\bibliographystyle{IEEEtran}
\bibliography{citations}

\end{document}